\documentclass[12pt,a4paper,oneside]{article}
\usepackage{amsmath,amsthm,amsfonts,amssymb,bm}
\usepackage{graphicx}
\usepackage{epstopdf}
\usepackage{enumerate}
\usepackage{hyperref}

\title{From CDF to PDF \\ A Density Estimation Method for High Dimensional Data}
\author{Shengdong Zhang \\ Simon Fraser University \\ sza75@sfu.ca}

\begin{document}
\maketitle

\newpage
\section*{Introduction}

Probability density estimation for high dimensional data is difficult. Kernel density estimation (KDE) for approximating probability density function (PDF) is a commonly used non-parametric technique which has been studied both empirically and theoretically.  However, this technique usually has poor performance when estimating probability density for data over $6$ dimensions, and suffers for finding optimal kernel functions and the corresponding bandwidths (i.e. smoothing parameters). In addition, data distribution tends to be sparser and sparser as the number of data dimensions increases, causing KDE, as well as other local methods, to give zero probability estimates for most of data space. Therefore, the need of data amount for the KDE method to give accurate density estimation grows exponentially with data dimension. Also, it does not scale well for "Big Data" due to its data storage requirement and computational burden for probability density inference. Similar issues go to other non-parametric PDF techniques such as K-nearest-neighbor.

PDF estimation based on Gaussian mixtures is one of many parametric techniques nicely studied. Its strong assumption (mixture of Gaussians) makes it difficult to estimate probability density for data sampled from highly non-Gaussian distribution (e.g. uniform distributions, highly skewed distributions) well. Similarly, all other parametric techniques also assume PDF of data has a certain functional form, and the functional form of the true PDF is usually unknown, causing most of the techniques rarely produce good estimated density for complex high dimensional data.

Estimating probability density using neural networks (NN) was proposed a long time ago. These methods usually give satisfying performance for low dimensional data, and become ineffective or computationally-infeasible for very 
high dimensional data. [1] and [2] propose a few methods to address this issue. [1] proposes two methods called Stochastic Learning of the Cumulative (SLC) and Smooth Interpolation of the Cumulative (SIC). Both of these methods make use of Cumulative Distribution Function (CDF). SLC only works for 1-dimensional data and SLC is able to deal with multidimensional data. But in the paper the authors only mention inferring PDF by differentiating the approximated CDF and no solution or algorithms for the computation of higher order derivatives provided. Such computation usually has no explicit formulas and hard to approximate numerically. [2] propose a method using a 3-layer neural net and training it with a modified loss function which includes a term approximating integration numerically over the region of interest, making the output of model valid probability estimates. However, the integration part of this method suffers from the difficulty of numerical integration for high dimensional data. 

Recent development of deep learning provides new algorithms for learning PDF by applying deep network structure. [3] discusses NADE and its deep extension DeepNADE, which make use of one-dimensional conditionals, to estimate probability density for multi-dimensional binary data . For real-valued data, its real value version RNADE requires a parametric functional form be selected for PDF estimation. [4] propose a PDF estimation algorithm called MADE, which combines conditional masking trick  inspired from the chain rule of probability and deep neural networks to enable a neural net model to learn valid probability estimates. But MADE is designed for binary data only and does not work for real valued data. 

CDF2PDF is a method of PDF estimation by approximating CDF. The original idea of it was previously proposed in [1] called SIC. However, SIC requires additional hyper-parameter tunning, and no algorithms for computing higher order derivative from a trained NN are provided in [1]. CDF2PDF improves SIC by avoiding the time-consuming hyper-parameter tuning part and enabling higher order derivative computation to be done in polynomial time. Experiments of this method for one-dimensional data shows promising results. 

\section*{Smooth Interpolation of the Cumulative (SIC)}

SIC[1] uses a multilayer neural network which is trained to output the estimates of CDF. Given $\vec{x} = [ x_1, \dots, x_d ]^T \in \mathbf{R}^d$, let $g(\vec{x})$  be the true PDF,  and $G(\vec{x})$ be the corresponding CDF defined as

\begin{center}
$G(\vec{x}) = \int_{-\infty}^{x_1}\cdots \int_{-\infty}^{x_d} g(\vec{t})dt_1\cdots dt_d$.
\end{center}

Given data $X = [\vec{x}_1, \dots, \vec{x}_d]$, to training a NN, $G(\vec{x})$ needs to be estimated from finite data samples. [1] provides two estimation methods for high dimensional data. One is to sample input space  uniformly and determine targets by the formula below:

\begin{center}
$\hat{G}(\vec{x})  = \frac{1}{N} \Sigma_{n=1}^N \Theta(\vec{x} - \vec{x}_n)$ \quad \quad $(1)$
\end{center}

\noindent where $\Theta(\vec{x}) = 1$ if $x_i >=0$ for all $i = 1, \dots, d$, and $0$ otherwise. Another option is to use the data points to generate targets by the formula similar to the previous one:

\begin{center}
$\hat{G}(\vec{x})  = \frac{1}{N-1} \Sigma_{n=1, n\neq m}^N \Theta(\vec{x}_m - \vec{x}_n)$ \quad \quad $(2)$
\end{center}

After generating training data by either one of these estimation methods, a multilayer neural network $H(\vec{x}, \vec{\theta})$, where $\vec{\theta}$ stands for model parameters, is trained with a loss function for regression. Since there are many choices for the loss function, we chose the square loss function in this work. 

Once training is done and $H(\vec{x}, \vec{\theta})$ approximates $G(\vec{x})$ well, the PDF can be directly inferred from $H(\vec{x}, \vec{\theta})$ by the formula:

\begin{center}
$\hat{g}(\vec{x})  = \frac{\partial^d }{\partial x_1, \dots, x_d }H(\vec{x}, \vec{\theta})$ \quad \quad $(3)$
\end{center}

\section*{Issues with SIC}

In [1], the following loss function is used for training.

\begin{center}
$L(\vec{\theta}, X) = \Sigma_{n=1}^N (H(\vec{x}_n) - \hat{G}(\vec{x}_n))^2 +\lambda\Sigma_{k=1}^{N}\theta(H(\vec{x_k})-H(\vec{x_k}+\Delta\vec{1}_d))[H(\vec{x_k})-H(\vec{x_k}+\Delta\vec{1}_d)]^2$ 
\end{center}

\noindent where $\lambda$ is a positive hyper-parameter that controls the penalty of non-monotonicity of the NN via the second term of the loss function, $\Delta$ is a small positive number, and $\vec{1}_d$ is a d-dimensional column vector of $1$'s.

Since $\lambda$ needs to be tuned, extra work such as cross-validation or grid-search are needed. Too large $\lambda$ could cause the trained NN to produce over-smoothed estimates and too small $\lambda$ cannot guarantee  the trained $H(\vec{x}, \vec{\theta})$ is non-decreasing over input space where contains very few or even no training data points, causing invalid probability density estimates later (i.e. negative values).

In addition, as stated before, no algorithm for computation of $(3)$ was proposed in [1]. The authors did  mention to estimate $(3)$ by numerical differentiation. However,  numerally approximating higher order derivatives is inefficient and numerically unstable, thus not feasible in practice.

\section*{Advantages of CDF2PDF Method}

In [1], the authors stated a few advantages of SIC:

\begin{itemize}
\item Approximating the CDF is less sensitive to statistical fluctuations than approximating the PDF directly from data, for the integral operator, or summation operation in practice, plays the role of a regularizer. 

\item Theoretically, the convergence rate of the PDF estimates inferred from the approximated CDF to a smooth density function which has bounded higher derivatives is faster than the convergence rate of KDE methods.

\item  Unlike KDE methods, there are no smoothing parameters that need to be determined for density estimation.

\end{itemize}

\noindent These advantages are also for CDF2PDF, and additionally it has more advantages below:

\begin{itemize}

\item CDF2PDF removes the non-monotonicity penalty term from the loss function and meanwhile guarantees the estimated CDF function is monotonically increasing and close to be monotonically non-decreasing. So the only hyper-parameter in this method is the number of hidden nodes.

\item  After the neural network learns the CDF from data, the corresponding PDF can be efficiently computed from model parameters in polynomial time.

\end{itemize}

\section*{CDF2PDF}

\subsection*{Neural Networks as Increasing Function Approximators}

In CDF2PDF method, only a single hidden layer neural net with sigmoid activation functions in the hidden layer and one linear output node is used for CDF estimation. A traditional single hidden layer neural net of this architecture has the following functional form:

\begin{center}
$H(\vec{x}) = \Sigma_{i=1}^H w_i^{(2)}\sigma(W_{i*}^{(1)}\vec{x} + b_i^{(1)}) +  b^{(2)}= \Sigma_{i=1}^H w_i^{(2)}\sigma(\Sigma_{j=1}^d W_{ij}^{(1)} x_j + b_i^{(1)}) + b^{(2)}$
\end{center}

\noindent where $\vec{x}$ is the input vector, $W_{i*}^{(1)}$ is the $i$-th row of $W^{(1)}$, H is the number of hidden nodes, $w^{(2)}, b^{(2)}$ are the weight vector and bias of the hidden layer, and $W^{(1)}, \vec{b}^{(1)}$ are the weight matrix and the bias vector of the input layer.

When the model parameters travel in the parameter space during training, values of $W^{(1)}$ and $w^{(2)}$ can be any real numbers. Previously, the second penalty term in the loss function $L(\vec{\theta}, X)$ for SIC is designed to regularize these values for making the trained neural net non-decreasing. In CDF2PDF, instead of using this penalty term, we directly enforce the neural net to be an increasing function. The functional form can be written as follows:

\begin{center}
$H(\vec{x}) = \Sigma_{i=1}^H e^{w_i^{(2)}}\sigma(\Sigma_{j=1}^d e^{W_{ij}^{(1)}} x_j + b_i^{(1)}) + b^{(2)}$
\end{center}

We called this type of neural networks Monotonic Increasing Neural Networks (MINN). The rational of this modification is that any composition function of increasing functions is again increasing. Knowing that any sigmoid functions are monotonic increasing, as long as parameter of $W^{(1)}$ and $w^{(2)}$ are positive, $H(\vec{x})$ must be an increasing function. Therefore, we replace  $W^{(1)}$ and $w^{(2)}$ with natural exponential functions of them. This modification removes the need of non-monotonicity penalty term in the loss function, and serve as a kind of regularization. Experiments below for $1$-dimensional data will show effectiveness of MINN's approximation ability of non-decreasing functions.

\subsection*{Inference of PDF}

After training is done, the corresponding estimated PDF can be inferred by:

\begin{align*}
\hat{g}(\vec{x}) &= \frac{\partial^d }{\partial x_1, \dots, \partial x_d }H(\vec{x})\\ 
 &=  \Sigma_{i=1}^H e^{w_i^{(2)}} \Pi_{j=1}^d e^{W_{ij}^{(1)}}\sigma^{(d)}(\Sigma_{j=1}^d e^{W_{ij}^{(1)}} x_j + b_i^{(1)}) \\
&=  \Sigma_{i=1}^H e^{w_i^{(2)} + \Sigma_{j=1}^d W_{ij}^{(1)}} \sigma^{(d)}(\Sigma_{j=1}^d e^{W_{ij}^{(1)}} x_j + b_i^{(1)})
\end{align*}

\noindent where $\sigma^{(n)}(x)$ is the $n$-th derivative of the sigmoid function. It indicates that to estimate a PDF of $d$-dimensional data from the MINN, the $d$-th derivative of the sigmoid function must be computed.

\subsection*{Computation of Higher Order Derivative}

Closed form computation for higher order derivative of sigmoid functions exists for its special differential structure.  For example, the second and third derivatives of tanh can be written as $\sigma ' (x) = 1 - \sigma^2 (x)$, $\sigma '' (x) = -2 \sigma (x)(1-\sigma^2 (x))$. Thus, the $n$-th derivative of the tanh function is a $(n+1)$-th degree polynomial of $\sigma (x)$. All we need is to determine the coefficients of the polynomial. An efficient algorithm is being developed and the mathematical details for it is included in [5]. 

\subsection*{Fine-Tuning for CDF of a Non-Smooth PDF}

In practice, not all CDF can be approximated well by smoothed functions such as sigmoid. It is hard for sigmoid functions to approximate strong non-smoothness, e.g. uniform distributions. To deal with this, after training a MINN, we replace all tanh activation functions with an modified version of tanh functions as follows:

\begin{center}
$\eta (x) = \rho  \sigma (x) + (1 - \rho)  s(x)$ \quad \quad $(4)$
\end{center}

\noindent where $\rho \in (0,1)$ and
\begin{center}
$s(x) = \begin{cases}
1 & \text{ if x$\geq 1$} \\ 
x & \text{ if $-1\leq x < 1$}\\ 
-1 & \text{ if $x \leq -1$}
\end{cases}$
\end{center}

For edge-cutting learning of $\rho$, we define $\rho = \frac{1}{1 + e^{-\alpha}}$, which $\alpha$ is a model parameter to be optimized during fine-tuning phrase.

Noticing that all higher order derivatives (second and any larger order) of $s(x)$ is $0$, such modification on the activation functions will not cause any issue for inferring high dimensional PDF.

\section*{Experiments}
\subsection*{Density Estimation for Multimodal Distributions}

We first test the proposed method for estimating the following density function called BartSimpson distribution.

\begin{center}
$f(x) = \frac{1}{2}N(x ; 0, 1) + \frac{1}{10}\Sigma_{j=0}^{4} N(x; (j/2)-1, 1/10)$
\end{center}

\noindent where $N(x ; \mu , \sigma)$ denotes a Normal density function with mean $\mu$ and standard deviation $\sigma$. $1000$ data points are sampled from $f(x)$. The following plots are histogram and empirical CDF of the sampled data.

\begin {figure}[h]
\centering
\includegraphics[scale=0.7]{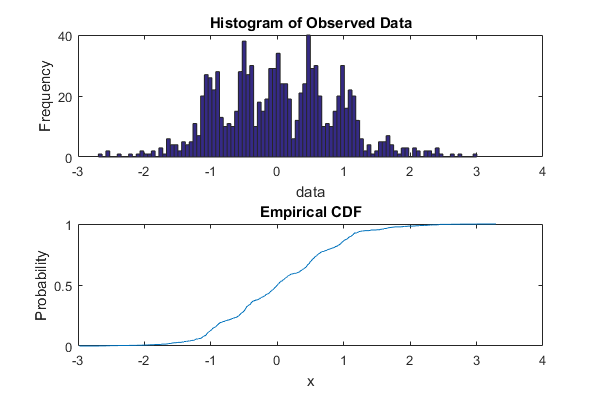}
\end{figure}

A MINN with $16$ hidden nodes is selected for training. Adadelta[6] is chosen as the stochastic optimization algorithm for training with batch size of $100$ and $30000$ epochs. The training data are generated by (1). Figure 1 shows approximated CDF vs emprical CDF,  and Figure 2 compares inferred PDFs from MINN and KDE with true PDF.

\begin{figure}[!tbp]
  \centering
  \begin{minipage}[b]{0.45\textwidth}
    \includegraphics[width=\textwidth]{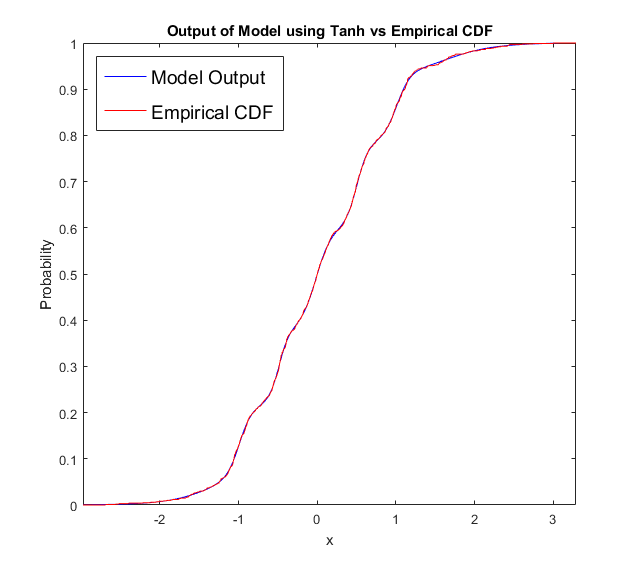}
   \caption{MINN Output vs Empirical CDF}
  \end{minipage}
  \hfill
  \begin{minipage}[b]{0.45\textwidth}
    \includegraphics[width=\textwidth]{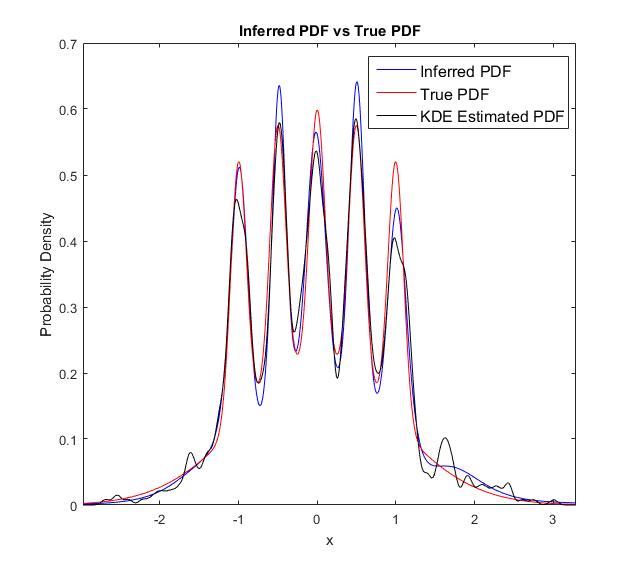}
  \caption{PDFs from MINN and KDE vs True PDF}
  \end{minipage}
\end{figure}


The PDF from KDE is estimated with Gaussian kernels and bandwidth $h = 0.05$ selected by cross-validation.  As we can see, CDF2PDF method is able to give better PDF estimates without any effort to hyper-parameter tuning, compared to the probability density estimates given by KDE.

Codes of this experiment correspond to \texttt{StochasticCDFLearning\_demo.m}  in \texttt{CDF2PDF\_1d\_codes} folder.

\subsection*{Density Estimation for Mixture of Distributions}
Next, we estimate the following density function of mixed distributions.

\begin{center}
$g(x) = 0.25N(x ; -7, 0.5) + 0.25U(x; -3, -1) + 0.25U(x; 1,3) + 0.25N(x; 7, 0.5)$
\end{center}

\noindent where $U(x; a, b)$ denotes a uniform distribution defined on interval $[a,b]$. $g(x)$ contains two uniform distributions, which make $g(x)$ not differentiable, thus not smooth, at boundaries of the uniforms. $2000$ data points are sampled from $g(x)$, and training data are generated by (1).

\begin {figure}[h]
\centering
\includegraphics[scale=0.7]{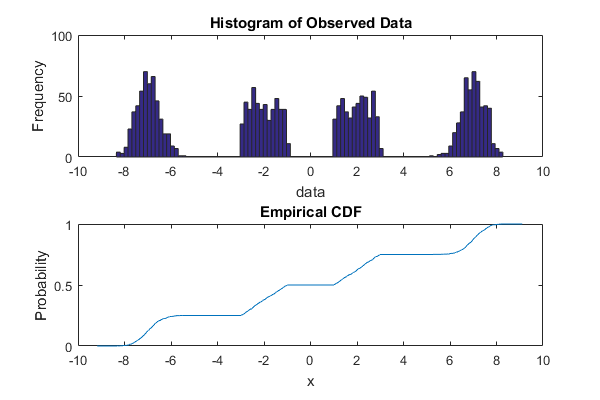}
\end{figure}

In this experiment, a MINN with $8$ hidden nodes is used. The optimization algorithm is again Adaddelta with batch size of $100$ and $10000$ epochs same as the previous experiment. Below are two plots showing approximated CDF vs emprical CDF and inferred PDF vs true PDF.

\begin{figure}[!tbp]
  \centering
  \begin{minipage}[b]{0.45\textwidth}
    \includegraphics[width=\textwidth]{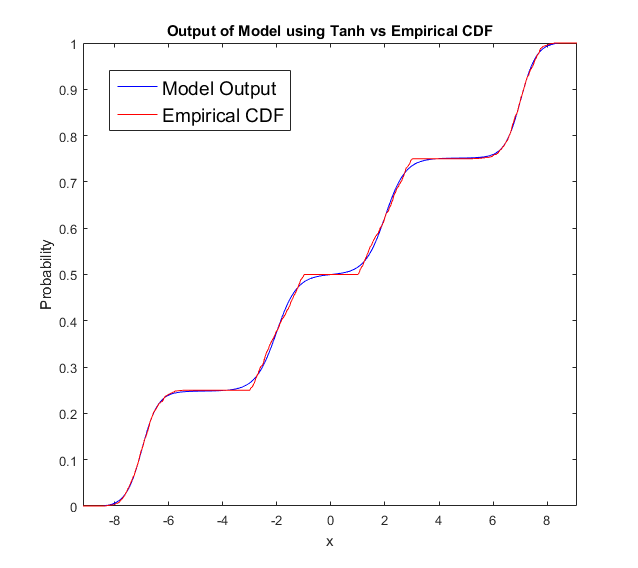}
  \caption{MINN Output vs Empirical CDF}
  \end{minipage}
  \hfill
  \begin{minipage}[b]{0.45\textwidth}
    \includegraphics[width=\textwidth]{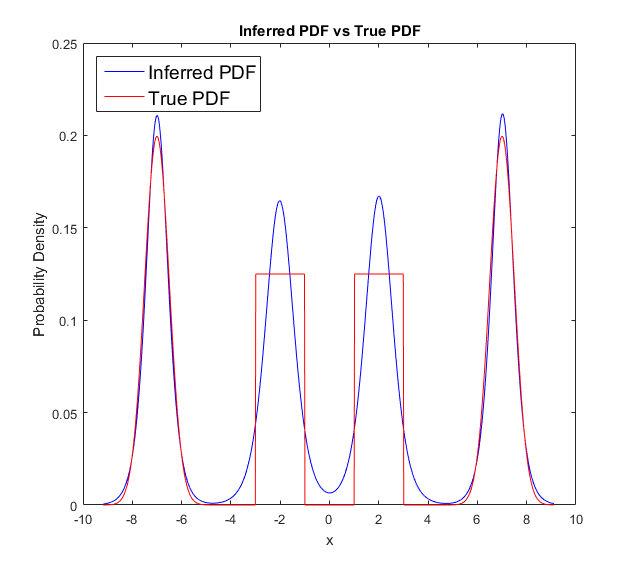}
  \caption{Inferred PDF from MINN with sigmoid functions vs True PDF}
  \end{minipage}
\end{figure}

To make the MINN adaptive to the non-smoothness of data, all activation functions in the trained MINN are replaced with $\eta(x)$ from $(4)$, and retrain the MINN along with $\alpha$'s in $\eta(x)$. The second training phase aims to fine tuning the nonlinear property of activation functions by $\alpha$'s. $5000$ epochs of training is performed for the fine tuning phrase.

\begin {figure}[h]
\centering
\includegraphics[scale=0.7]{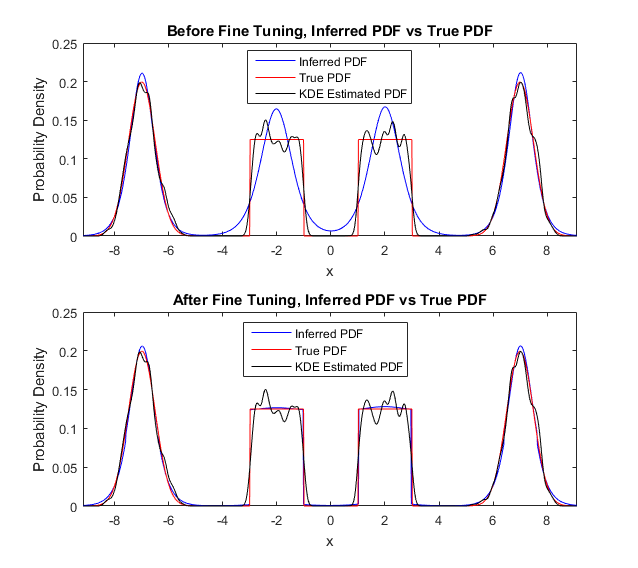}
\caption{PDFs of MINN before and after Fine Tuning vs True PDF}
\end{figure}

Figure 5 shows the effectiveness of the fine-tuning phrase. In Figure 5, the PDF estimates given by KDE uses Gaussian kernels with bandwidth $h = 0.1$ determined by cross-validation. The PDF estimates of MINN after fine-tuning is much closer to the true PDF and less fluctuate over the intervals of the two uniforms compared to the KDE estimates.

Codes of this experiment correspond to \texttt{FineTuning4CDFLearning\_demo.m}  in \texttt{CDF2PDF\_1d\_codes} folder.

\section*{Discussion}
Probability density estimation is fundamental and essential in most fields of science and engineering. Directly estimating the density function from data is an ill-posed problem and naturally hard for high dimensional data due to "the curse of dimensionality". Based on the experimental results for $1$-dimensional data, it seems promising to extend the CDF2PDF method to high dimensional data. Potential solutions to the major issues for computing the estimated high dimensional PDF from a trained MINN are given in this article. For theoretical discussion about convergence of estimated PDF from CDF to the true PDF,  please refer to [1].

\section*{Current \& Future Work}

Currently, I am working on developing efficient algorithms for fast generation of training data and higher order differentiation. The CDF2PDF method proposed in this article is only a prototype. It is definitely possible to build a deep MINN while keeping higher order derivative computation feasible, which belongs to future work.

\section*{References}

\begin{enumerate}[ {[}1{]} ]
\item Magdon-Ismail, M., Atiya, A. (2002). Density Estimation and Random Variate Generation using Multilayer Networks. {\it IEEE Transactions on Neural Networks}, 13(3), 497-520.
\item Likas, A. (2001). Probability Density Estimation using Artificial Neural Networks. {\it Computer physics communications}, 135(2), 167-175.
\item Uria, B., Côté, M. A., Gregor, K., Murray, I., Larochelle, H. (2016). Neural Autoregressive Distribution Estimation. {\it Journal of Machine Learning Research}, 17(205), 1-37.
\item Germain, M., Gregor, K., Murray, I., Larochelle, H. (2015, February). MADE: Masked Autoencoder for Distribution Estimation. In {\it ICML} (pp. 881-889).
\item Boyadzhiev, K. N. (2009). Derivative Polynomials for Tanh, Tan, Sech and Sec in Explicit form. arXiv preprint arXiv:0903.0117.

\end{enumerate}

\end{document}